# Tiling and stitching segmentation output for remote sensing: basic challenges and recommendations

*Bohao Huang[1], Daniel Reichman[1], Leslie M. Collins[1], Kyle Bradbury[2], and Jordan M. Malof[1]*

[1]Department of Electrical & Computer Engineering, Duke University, Durham, NC 27708
[2]Energy Initiative, Duke University, Durham, NC 27708

*Abstract*— In this work we consider the application of convolutional neural networks (CNNs) for pixel-wise labeling (a.k.a., semantic segmentation) of remote sensing imagery (e.g., aerial color or hyperspectral imagery). Remote sensing imagery is usually stored in the form of very large images, referred to as "tiles", which are too large to be segmented directly using most CNNs and their associated hardware. As a result, during label inference, smaller sub-images, called "patches", are processed individually and then "stitched" back together to create a tile-sized label map. There are many variants of stitching in the literature involving, for example, averaging overlapping labels, or clipping labels near the edges of the output label image. There is relatively little explanation or justification offered for these variants in the literature, and little experimental evidence of the impact or superiority of any particular approach. To address these limitations, we provide a survey of existing stitching approaches, and then explain how all approaches are fundamentally motivated by *translational variance* of segmentation networks – that is, the label predicted for a particular pixel depends upon its relative position in the input patch. We explore the primary causes of translational variance in modern CNNs, and support this with experimental evidence. Finally, we recommend a stitching strategy to maximize label accuracy and minimize computational costs. The proposed method contributed to our winning entry in the INRIA building labeling competition.

*Index Terms*— semantic segmentation, convolutional neural networks, deep learning, aerial imagery, building detection

## I. INTRODUCTION

Convolutional neural networks (CNNs) are now the dominant method for semantic segmentation (i.e., dense pixel-wise labeling) of remote sensing imagery, such as color or hyperspectral satellite imagery [1]–[8]. For example, performance in several recent benchmark problems has been dominated by CNNs including a recent Kaggle competition for building labeling [5], the INRIA building labeling competition [7], and the recent ISPRS labeling competition [8].

Here we consider the unique challenges of performing label inference on large remote sensing imagery using semantic segmentation CNNs, termed segmentation networks (SNs). Raw remote sensing imagery is often stored as large image "tiles", which cannot be processed directly, as a whole, because of limited memory on the graphics processing units (GPUs) used by modern CNNs. A common solution to this problem is to extract smaller sub-images, termed *patches*, and process them individually. This process – termed *stitching* - is illustrated in Fig. 1. Once label patches are inferred for each input patch, they are placed back into position in order to form a label tile.

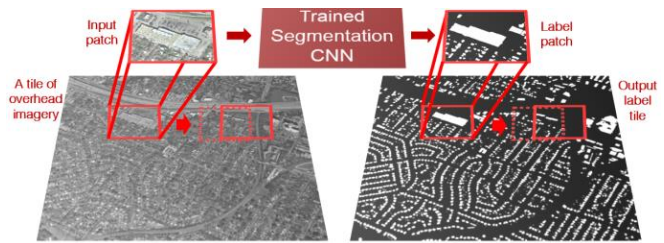

Fig. 1. Illustration of stitching for building segmentation in color overhead imagery. Small sub-images, termed *patches*, of imagery are extracted from large tiles of remote sensing imagery (left). Each patch is processed individually, producing a label patch (top right). The output label patches are then placed back into position in the tile, or "stitched", forming a large contiguous label tile (bottom right). "Stitching" includes many variations such as densely extracting label patches so that the overlapping labels can be averaged; clipping the edges of label patches; or simply concatenating neighboring label patches. The goal of this work is to elucidate the motivation for these different designs, and understand which are best.

### A. Variations of stitching

Label stitching is not simply the process of concatenating label patches however, and there are three major variants in the literature. These variants employ additional processing with the goal of improving label accuracy. Perhaps the most common class of approaches involves overlapping the output label images so that the labels overlap and can be averaged [2], [9]–[12]. The amount of overlap varies across approaches, resulting in different numbers of labels being involved in the averages. One study also applied a weighted averaging scheme among overlapping pixels [9].

A second class of approaches involves clipping the edges of the output label patches, and then concatenating the remainder of the patch without overlap or averaging [5], [7], [13]. A final class of approaches simply concatenate the output label patches without any modification [3].

We provide additional explanation and motivation for these approaches in Section IV. For now, it suffices to say that, despite the existence of several classes of stitching approaches, little justification or experimental evidence has been provided for any particular approach. In general, it is unclear why one approach may be preferable.

### B. Contributions of this work

In this work, we aim to elucidate the process of stitching, and provide guidance on how to achieve the best tradeoffs between computational costs and label accuracy. Towards this goal, we analyzed existing stitching approaches and their motivations. A central finding of that investigation is that all existing methods are motivated by the absence of perfect *translational*

*equivariance* in modern CNNs. A network is translationally equivariant if a translation of the input patch results in a translation of the corresponding label patch, without any other changes in the predicted labels. However, modern CNNs are not often perfectly equivariant, and yield changes to their label predictions after [14].

We use the term *translational variance* to refer to the variability of the predicted labels with respect to translation. This behavior is illustrated in Fig. 2. In this work we identify the underlying causes of *translational variance* in many modern segmentation networks: non-unary strides, and zero-paddings. We support these assertions with mathematical and experimental evidence, and we quantify the relative impact of each cause on label accuracy.

Based on our findings we recommend a stitching strategy to maximize label accuracy and minimize computational time – the major goal of this work. We support our recommendations with experimental results to illustrate the tradeoffs between label accuracy and computation. The proposed stitching approach was employed in our winning entry to the recent INRIA building labeling competition [7], [15], and is an extension of preliminary work [13].

### C. Paper organization

The remainder of this work is organized as follows: Section II and Section IV discuss the remote sensing datasets and the segmentation networks considered in our experiments, respectively. Section IV reviews existing motivations for stitching approaches, and discusses the central role of translational variance. Section V discusses the two major causes of translational variance, including theory and experimental evidence to support the discussion. In Section VI we recommend a general stitching approach, and provide experiments to support its advantages. In Section VII we present conclusions.

## II. THE SEGMENTATION DATASETS

To evaluate our hypotheses in this work we run experiments on two large aerial imagery datasets. One of them is The INRIA building labeling dataset (D1) [15] and another one is the solar array labeling dataset (D2) [16].

### A. The INRIA building labeling dataset (D1)

Dataset 1 (D1) is the INRIA Aerial Image Labeling Challenge Dataset [15]. We selected D1 because it is a popular benchmark dataset, and because its geographic diversity will include many different stitching conditions: different shapes and sizes of objects, and different class priors. This dataset contains aerial RGB imagery collected from 10 cities in both the U.S. and Europe, however in this work, we only used the 5 cities with publicly available ground truth labeling: Austin, Chicago, Kitsap, Western Tyrol, and Vienna. A total of 36 images were captured over each city at a ground sampling rate of 0.3 m. Each of the 36 images encompasses 2.25 km$^2$, which translates to 5000 × 5000 pixels in each tile.

### B. The solar array labeling dataset (D2)

Dataset 2 (D2) is a color (RGB) dataset of ortho-rectified aerial photography for the problem of pixel-wise solar photovoltaic array labeling [16]. D2 was included principally to help quantify the variability of our results across different remote sensing problems. Covered more than twice of the area as D1, D2 also has the advantage of abundant data. We used a subset of the data comprising roughly 19,000 solar arrays, over 1000 km$^2$ of area collected over three municipalities in California, U.S.A: Fresno, Modesto, and Stockton. This subset was chosen because all of the imagery was collected at the same 0.3-meter resolution. This dataset has been employed in several studies of semantic segmentation [9], [17].

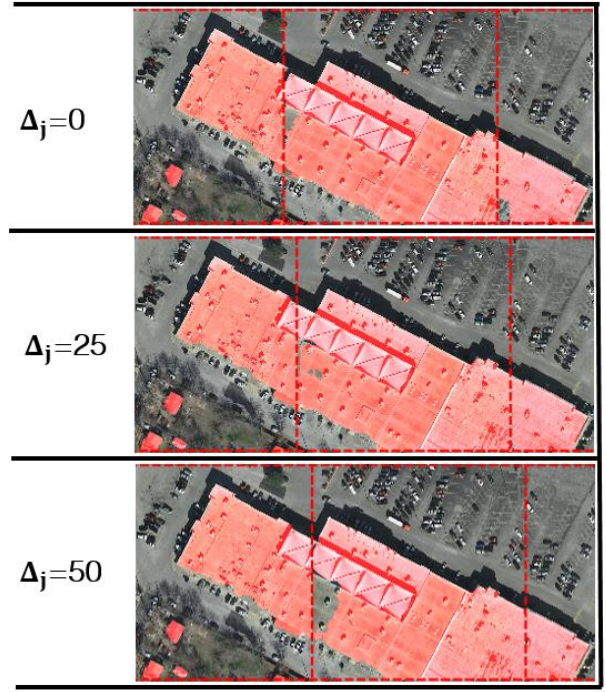

Fig. 2. Illustration of translational *variance* in the label patches predicted by a modern segmentation network – the U-Net. Each row shows the same (color) image patches, with the predicted labels from a segmentation network overlaid in transparent red. The boundaries between neighboring label patches are indicated by red dashed lines. The term $\Delta_j$ indicates how much the imagery is horizontally translated with respect to the CNN. Each row shows a different level of translation. Horizontal translations manifest themselves as horizontally translated dashed lines in the borders because the area of the input imagery is held constant – this causes it to appear as though the network is translating. This was done deliberately so that each input pixel does not move in the visualization, and therefore it is easy to compare the predicted labels across different translations. We see that the same pixel receives a different label, depending upon $\Delta_j$. This often manifests itself as discontinuities at the borders of neighboring label patches.

## III. SEGMENTATION NETWORKS

In this work we used two popular semantic segmentation models: U-Net [18] and DeepLabV2 [19]. We select these two models because (i) they achieved state-of-art performance in several benchmark datasets [8], [15] and (ii) they represent two popular design choices for segmentation CNNs, making our findings more relevant. The U-Net uses an encoder-decoder structure [7], [20], while the DeepLab model uses a ResNet encoder (or "backbone") with Spatial-Pyramid-Pooling [21]–[24]. We will see that these two approaches also result in somewhat different behaviors with respect to stitching, making it important to analyze both of them. We briefly review these architectures here.

## A. The U-Net architecture

The U-Net is a popular semantic segmentation CNN architecture [18] that was originally proposed for the segmentation of medical imagery [18]. We use the U-Net architecture as it was proposed in [18], with the single exception that we use half as many filters in each convolutional layer. This modification was adopted because it was used by the winning entry in the INRIA building labeling competition [15]. Note that the U-Net model does not employ any zero-padding in its intermediate convolutional (feature) layers. Therefore, its output feature maps (and final label maps) are smaller than its input.

## B. The DeepLabV2 architecture

The DeepLabV2 models adapt Atrous Convolution to maintain the spatial extent of the output feature maps throughout convolutional layers. In [19], the DeepLabV2 implementation based on the ResNet-101 architecture [25] was reported to outperform the implementation based on VGG16 [26] for the PASCAL-VOC 2012 dataset [27]. Therefore, we chose to utilize the DeepLabV2 implementation with the ResNet-101 network architecture in our experiments. Since we are focusing on addressing stitching problems specifically associated with CNNs, we did not include the Conditional Random Field (CRF) often applied after post processing.

Note that, in contrast to the U-Net model, the DeepLabV2 model uses zero-padding throughout its intermediate convolutional (feature) layers in order to help maintain the spatial extent of its output label patches. As we hypothesize in Section V, this has the effect of reducing the accuracy of the label predictions of the DeeplabV2 at the edges of its output label maps.

## C. Network training

We trained both of the CNN models (DeepLabV2 and U-Net using D1 and D2. For D1, the first five tiles in each city formed the validation set and the remaining 31 tiles in each city were used for training. For D2, we used the first half of the images in each city as the validation set, and the used the remaining half as the training set.

The optimization procedure and the related parameter settings in all of the experiments are consistent across models. The optimization objective function is the discrete cross entropy loss, which is widely used [28]. Unless specified, we use a batch size of 5 and patch size of $572 \times 572$ pixels for the U-Net models (with no zero padding) and $321 \times 321$ pixels for the DeepLabV2 models as the default choice of their original implementations. An Adam optimizer [29] with $\beta_1 = 0.9$, $\beta_2 = 0.999$, $\epsilon = 10^{-8}$ is used. The models are trained for 100 epochs with 8,000 patches per epoch. We did a grid search of hyper-parameters and select corresponding parameters that yield best results. For the experiments with the U-Net no zero-padding model, we trained the networks with a learning rate of $10^{-4}$ and dropped to $10^{-5}$ after 60 epochs. For the experiments with the DeepLabV2 model, we trained them with a learning rate of $10^{-5}$ and dropped to $10^{-6}$ after 60 epochs.

## IV. MOTIVATIONS FOR STITCHING AND THE ROLE OF TRANSLATIONAL VARIANCE

In this section we discuss motivations for the three major stitching approaches in the literature: label averaging, label clipping, and concatenation. For each approach we explain the major motivations for it that have been provided in the literature, and any theoretical or experimental support that has been provided. In all cases, we also explain how each of these approaches make some assumptions about translational variance in segmentation networks.

## A. Label clipping

These approaches involve clipping (or removing) labels at the edge of the output label patches. This is based on the notion that these edge labels exhibit relatively high error rates, on average, compared to labels near the center of the label patches. If edge labels do indeed exhibit higher error rates, then it is beneficial (for label accuracy) to clip the label-image edges to remove lower-accuracy labels. We will see in IV.B that this is also sometimes cited as a justification for averaging overlapping labels as well.

Evidence for this notion has been provided in terms of zero-padding in one work, and the authors in [5] showed experimentally that label errors did indeed increase towards the edges of the output label image for their network (a U-Net). We reproduce these results on another dataset, and with an additional modern segmentation network (DeepLabV2).

The notion that edge labels exhibit greater error rates implies that any pixel, if it happens to reside at the edge of an input patch, may receive a worse prediction than if it happened to reside towards the center. This is a special case of translational variance because it implies that any pixel, once moved to the edge of an input patch via a translation, will occasionally receive a worse (i.e., different) label. We assume here that imagery at the edges of the input patches is not inherently more challenging to label – an alternative explanation.

## B. Label averaging

This approach involves extracting patches densely, so that the output label patches overlap. The overlapping labels are then averaged in order to improve their accuracy. This approach implicitly assumes that overlapping (i.e., spatially coincident) labels will occasionally disagree, since there is no benefit to averaging labels that are always identical. However, the only major difference between any two labels that overlap is that the input imagery has been translated with respect to the segmentation network. In turn, this implies that spatially coincident label predictions will occasionally disagree due only to translation, which is an implication of translational variance.

Although averaging labels assumes the presence of translational variance, authors have implicated different forms of translational variance. In many cases authors have suggested that edge labels exhibit higher error rates, and that these error rates can be reduced through averaging [10]–[12]. This is the same motivation often given for edge clipping, but with an alternative solution. Relatively little experimental evidence has been provided to support averaging edge labels, except in [10] it was reported that it provided a 1% increase in overall label accuracy. As we discuss in Section VI, averaging may be beneficial, but clipping is probably a superior approach.

In other studies, no particular type of translational variance was implicated, but some type of label averaging was employed. For example, [4], [6], [11] employed label averaging without citing any particular motivation. Several papers have also employed label averaging so that it involves averaging labels that are not necessarily near the edges of the imagery [10]–[12], suggesting that the authors believe there are indeed other causes of translational variance, that may introduce variance across the entire label map. As we will show in Section V, such translational variance does exist, although it is much less impactful.

*C. Label concatenation*

The last class of approaches simply involves concatenating the output label patches, without any clipping or averaging [3]. This approach implies that averaging or clipping would not be beneficial. This could be because the labels simply do not vary with respect to translations (i.e., the networks are translationally equivariant) or that the translational variance exists, but it is inconsequential enough that averaging or clipping would not be beneficial. As we discuss in Section V, translational variance of at least two kinds exists in CNNs, and ignoring it could lead to lower label accuracy.

## V. COMMON CAUSES OF TRANSLATIONAL VARIANCE

In this section we explore two major causes of translational variance in modern segmentation networks: zero-padding and max-pooling. We focus on these two causes of translational variance because they are widely used operations in modern segmentation networks, and as we show, they explain the translational variance motivating all variants of stitching. As we describe subsequently, zero-padding results in poorer label accuracy at the edges of label patches. Max-pooling does not lower label accuracy in any general way, but it introduces translational variance in the network that can be reduced via averaging and improve label accuracy.

While both operations give rise to translational variance, their impact on label accuracy is very different, and the computational costs of addressing them are different. This makes it useful to disentangle and address them separately. For both zero-padding and max-pooling we explain quantifying their relationship to translational variance and support our assertions with experiments.

*A. Zero-padding*

Zero-padding is a common method used in convolutional operations (e.g., filtering), and within CNNs. Zero-padding involves adding zero-valued pixels around the perimeter of an image (or other data), with the goal of maintaining the spatial or temporal size of the data after processing with convolution, max-pooling, and other common CNN operations. Padding is illustrated in Fig. 3 for convolution with a 3×3 pixel receptive field (i.e., input size).

The plausible problem with zero-padding is that it introduces zeros rather than real data into processing, potentially yielding lower-quality output. In CNNs, this output usually also serves as input to subsequent layers, propagating the errors over a wider spatial area due to the non-unary receptive fields of subsequent layers. This propagation of error is illustrated in Fig. 3.

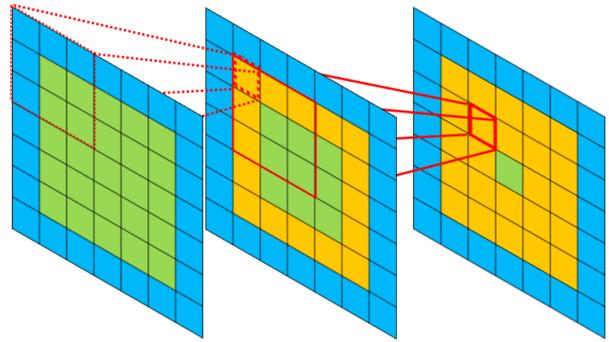

Fig. 3. An example of how errors caused by zero-padding propagate through convolutional layers. Original pixels are in green cells, zero padded pixels are in blue and affected pixels are marked in yellow. Passing a 5 by 5 pixel data cube through two 3 by 3 convolutional layers will result in only the center pixel being unaffected.

The precise impact of zero-padding will depend upon the particular statistical characteristics of the data, and therefore will vary. Fig. 4 quantifies the impact of zero-padding on one of our datasets (D1) for U-Net. We use a controlled experiment in which we apply U-Net with, and without, zero-padding. The results indicate that, when zero-padding is present, it increases the error label rates by as much as 35% at the edges. A similar result was reported in [5] for the U-Net model but with a different dataset. Thus are results here provide further support for this finding. Here we also present the error rate of the DeepLabV2 model. Due to the number of layers, it is infeasible to remove zero-padding from the model, and so there is no control, but it still appears that the label error rates rise towards the edge of the label patch.

Note that this zero-padding effect is equivalent to the "lack of context" some authors have cited in the literature for less-accurate edge labels [5]. That is, predicted labels at the edge of the label patch were based upon less data than those towards the center, by virtue of being at the edge. However, if you remove all zero-padding, then the label patch becomes smaller (sometimes substantially) than the input patch, and every label predicted by the network is based upon a complete set of real data. This is precisely what happens in the U-Net, resulting in relatively uniform label accuracy in Fig. 4.

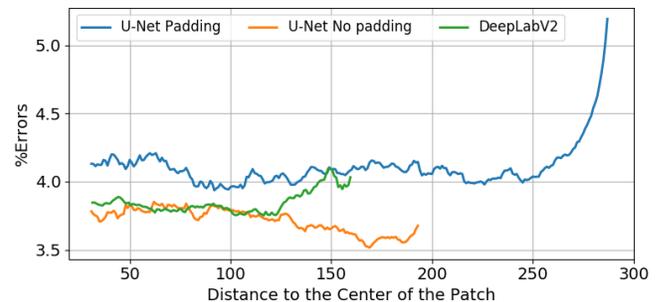

Fig. 4: Percentage of errors across patches for U-Net and DeeplabV2 CNN models when applied to our INRIA dataset (D2). The length of each curve is determined by its input patch size. The distance in the x-axis represents $\max(l_h, l_v)$ where $l_h, l_v$ are the corresponding horizontal and vertical distances in the Manhattan-distance. We are excluding the pixels with distance smaller than 30 pixels because with total number of patches fixed, there are only few pixels within this range and the results are therefore noisy.

## B. Non-unary stride

Perhaps the primary cause of translational variance in CNNs are non-unary strides (i.e., greater than one) in the network layers. The term *stride* refers to the pixel-wise step size that convolutional operations, such as pooling or filtering, take across an input image. We will denote the stride of a convolutional operation with the symbol $\delta$. A non-unary stride is one in which $\delta > 1$, in which case the output image is smaller by a factor of $\delta$.

Non-unary strides introduce translational variance because they cause the image output to be smaller than its input by a factor of $\delta$. This decimation of the input image causes unary translations of the input image to result in fractional (sub-unary) translations in the output image. The result of these fractional strides is that the output labels simply change value, rather than translating. This effect is illustrated in Fig. 5 for the popular max-pooling operation (e.g., the U-Net and DeepLabV2 considered in this work both use max-pooling operations).

Let $I$ denote a 1-dimensional input image that we will input into a single-layer network. The network is comprised of a convolutional operation acting on the input image, $f(\cdot)$, with a stride of $\delta_0$. We have $I' \coloneqq f(I)$, where $I'$ is the processed image. The image has a new coordinate system that is related to the old coordinate system by

$$j' = g(j) = j/\delta \quad \forall k' \in \mathbb{N} \quad (1)$$

The implication here is that translating the input image by a value, $k < \delta$ results in a fractional translation of the output image. Let $\delta = 2$, then $j' + \frac{1}{2} = g(j+1; \delta = 2)$. This translation causes us to sample $I'$ at a fractional location, which has different values than the output if no translation is applied. Alternatively, $j' + 1 = g(j+2; \delta = 2)$, in which case $I'$ exhibits the same values, but translated by one. Both of these scenarios are illustrated in Fig. 5 for the max-pooling operation.

The aforementioned observation implies that a network is translationally *equivariant* with respect to input-image translations that are multiples of $\delta$. We specified no particular form for the function, $f(\cdot)$, and therefore this result applies to any common convolutional operation that can have non-unary strides, such as regular, Atrous [19], and transpose convolutional or upsampling layers; or max, average, or learned pooling layers.

This result generalizes to larger networks that involve several successive pooling operations that may be interspersed with other layers. If $N$ denotes the number of pooling layers in a network, each with stride $\delta_i$, where $i^{th}$ non-unary-strided operation, then we have

$$\delta_{tot} = \prod^N \delta_i \quad (2)$$

where $\delta_{tot}$ is the effective stride between the input and output of a multi-layer network. If the individual strides are all the same value, then $\delta_{tot} = \delta^N$. We provide a sketch of the proof for this assertion in Appendix I.

To demonstrate the relationship between pooling and translational equivariance, we conduct an experiment with the DeepLabV2 and U-Net segmentation networks. The experiment is designed to measure the correlation between the labels predicted for a single pixel, as the pixel is translated with respect to the segmentation network. We expect that the labels will have correlations of approximately one (exact match) when we translate the imagery by factors of $\delta$.

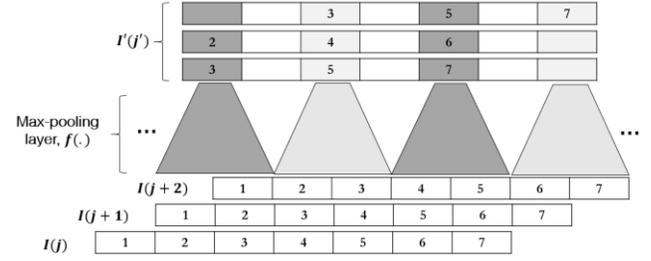

Fig. 5. Illustration of a convolutional max-pooling layer, with a stride $\delta = 2$, operating on a 1-dimensional input image, $I$. This operation returns the maximum value of the two input values. The output of the max-pooling layer is shown with respect to three different translations of the input. The output image, $I'(j')$, is presented for each possible translation. A translation of one in the input equates to a translation of ½ in the output, which manifests itself as a change in the output, and no translation. This is a form of translational variance in the network. Notice however, that if we translate by two pixels, the original output has simply been translated. This indicates that the network is *equivariant* with respect to input translations that are multiples of $\delta$.

To conduct this experiment, we select four 100x100 pixel regions from each tile in dataset one (D1). Then we record the labels predicted for each pixel in each 100x100 pixel region as we shift the input patch by $\Delta_j, \Delta_i$ pixels in both the horizontal and vertical axis, respectively while maintaining that the patch covers the 100x100 pixel region. The labels from each translation are used to obtain a correlation matrix for U-Net and DeeplabV2, presented in Fig. 6.

The U-Net has 4 max-pooling layers, each with $\delta = 2$. Furthermore, the U-Net has no zero-padding that may also introduce translational variance. Therefore we expect translations that are multiples of $2^4 = 16$ pixels to yield a perfect correlation, and indeed that is the case for the U-Net. The DeepLabV2 is based on the ResNet-101 architecture, which has three layers with $\delta = 2$, and we see that the correlations have peaks at translations of $\delta^3 = 8$. However, the correlations never reach a value of one, we hypothesize due to the substantial zero-padding, which introduces additional translational variance at across all translations. We can see also that the peaks fall farther from a value of one as the total translation size increases (i.e., pixels move closer to the edge), which is consistent with this hypothesis.

## C. Trans. variance is only caused by two sources

We want to draw additional attention to the correlation results with the U-Net in Fig. 6. The fact that the U-Net achieves full translational *equivariance* - after accounting for non-unary strides – suggests that the only two apparent sources of translational variance in the U-Net are (i) zero-padding, if it is used, and (ii) non-unary strides. Since most networks are comprised of the same components as the U-Net, this is a fairly general statement about the sources of translational variance in CNNs.

## VI. LABEL STITCHING RECOMMENDATIONS

In this section we provide recommendations for stitching based on the discussions in Section V and VI. The two main performance metrics we consider are computational time and label accuracy, in terms of the popular intersection-over-union (IoU) metric [27]. We make three primary recommendations, each described in its own subsection.

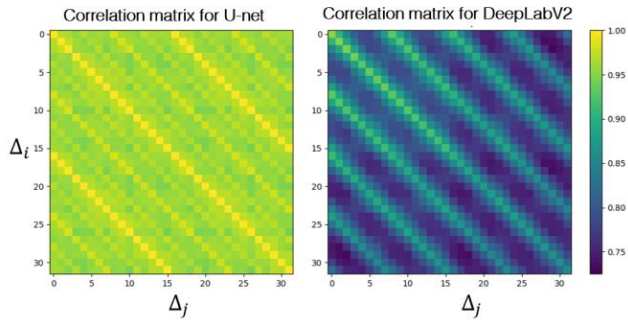

Fig. 6. The correlation matrices of the U-Net and the DeepLabV2 models on D1. We translated images by 32 pixels both horizontally and vertically – one pixel at a time – and then calculated the correlation of the same center 100 by 100 pixel region. Each row shows the correlation between horizontal translations, $\Delta_j$, and columns shows the correlation between vertical translations, $\Delta_i$. The U-Net models has high correlations globally and also a value of 1 every 16 pixels. The DeepLabV2 shows peak correlations every 8 pixels, however they are not quite equal to 1. Also the peaks become smaller as the total translation increases. See the text for explanations of these phenomenon.

### A. Increase the output size during label inference

We first make a novel strategy for stitching, which involves increasing the input-image size of the segmentation network *only during label inference*. Because modern networks are usually fully convolutional, their input size can be altered at any time, including for label inference, after the network has already been trained. We recommend the input size to the maximum memory capacity of the hardware (usually the RAM of the GPU). We find that this allows us to increase our input imagery from roughly 500x500 pixels to 2500x2500 pixels.

This approach has several advantages. First, it minimizes the amount of pixels in the output label patch that are impacted by zero-padding. As we show, this can improve label accuracy and is compatible with other processing such as label clipping, averaging, or doing nothing. Another advantage is reduced computational time. Much of the computational time required for modern CNNs is comprised of data handling (e.g., passing data between the GPU and the CNN, subsampling and stitching patches, etc.). Using this approach reduces this overhead, and results in substantially lower computational time. Next we conduct experiments to demonstrate the impact of this approach on computational time and label accuracy (IoU).

*1) Computational time improvements*

The computational time for generating label predictions with the U-Net and the DeepLabV2 models is shown in Fig. 7. The trend in performance is similar for both of the models that were considered: as the patch size increases, the running time decreases. The running time decreases due to (i) reduced data handling overhead because fewer forward network passes are needed and (ii) because less stitching operations are required. The simplicity of this approach and its faster running time are likely its greatest advantages.

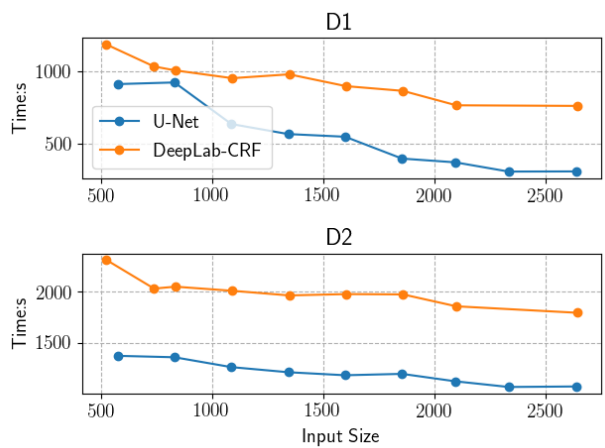

Fig. 7. Running time comparison for the U-Net and DeepLabV2 generating label predictions on all of the images in the validation data set in D1 and D2. In each plot, the input size at testing is shown on the X-axis.

*2) Label accuracy comparisons*

In Fig. 8 we compare the segmentation performance of the U-Net and DeepLabV2 models by measuring the IoU on validation images. For the DeepLabV2 model, there is a large performance gain when increasing the input size at testing. As shown in Fig. 4, the DeepLabV2 model generates poorer results at the boundary of input patch. By increasing the input patch size, the percentage of the output pixels impacted by this effect can be reduced, resulting in an overall performance improvement in the stitched output label maps.

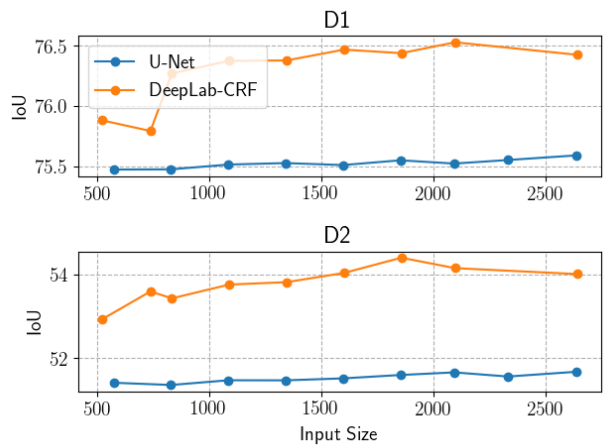

Fig. 8. IoU of U-Net and DeepLabV2 on the validation data set from D1 and D2. In each plot, the input size at testing is shown on the X-axis.

Another more surprising experimental result is the small, but consistent, performance improvement the U-Net model as the patch size increases. We see in Fig. 4 that the U-Net has similar errors across its output label patch, and therefore a performance improvement due to label clipping is surprising. We hypothesize that there may still be relatively higher error rates at the edges of its output, even though they are not apparent in Fig. 4 once zero-padding is removed.

## B. There is little benefit to averaging labels

As we noted in Section II.C, a popular stitching strategy involves averaging the labels from overlapping label patches. As discussed in Section V, the only reason to average labels (in the manner that others have done) is if they vary due to translation. As we discussed in Section VI.C, once zero-padding is removed from networks, the only likely source of translational variance is due to non-unary strides within the network layers. Therefore an important question is whether it is useful to reduce this variance through averaging.

To examine this hypothesis, we input translated versions of the input imagery into the U-Net and DeepLabV2 models, averaged all overlapping labels, and then evaluated whether IoU improves. We perform this experiment on both of our experimental datasets, and the results are presented in Fig. 9 below. Note that these experiments are performed using the strategy recommended in VII.A of increasing the input-patch size, and therefore the impact of zero-padding is substantially reduced in the DeepLabV2, and it was already absent from the U-Net.

The experimental results indicate that the U-Net receives little or no performance improvement from averaging, indicating that averaging out the translational variance from non-unary strides is not very beneficial. The DeepLabV2 includes additional translational variance due to the zero-padding, which we established is detrimental to label accuracy (see Fig. 4). As we'd expect therefore, the DeepLabV2 model does exhibit slightly greater performance improvements from averaging, and it does so consistently, yielding improvements over both datasets. However, once again, the performance improvements are modest, despite requiring 15x more processing because 15 translated version of the input imagery must be processed.

Although the computation-versus-performance tradeoffs are unfavorable with averaging, there appears to be little risk to this operation (i.e., of lowering performance), and therefore, when label accuracy is important, it appears to be somewhat beneficial.

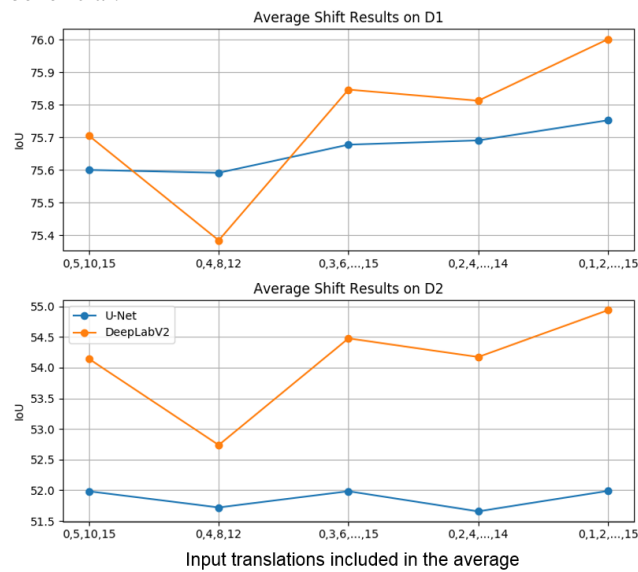

Fig. 9. IoU performance comparison for both the U-Net and the DeepLabV2 model on D1. The number of shifted pixels being aggregated is labeled in the X axis.

## C. Clip label edges to maximize accuracy

In Section V.A we showed that clipping the edges can mitigate label errors. Clipping the edge labels is intended to remove the labels that are most negatively impacted by zero-padding. With the U-Net, or other relatively shallow networks, this can be done simply by not zero-padding convolutional layers of the network. This shrinks the output of the network somewhat, but if the network is shallow enough, it doesn't introduce substantial computational overhead: this is especially true given our recommendations in Section VII.A to increase the input size, in which case clipping has minimal impact on processing time, while maximizing performance.

Our recommendations are a little more complicated for deep networks (i.e., those with many layers), such as DeepLabV2, in which there are very many layers with zero-padding. In these cases, many labels will be contaminated by zero-padding. For example, DeepLabV2 is based on the ResNet-101, and therefore all predicted labels will involve some zero-values.

Furthermore, labels that are closer to the edge will have relied upon more zero-values, and therefore they might exhibit lower accuracy. The results of Fig. 4 imply that there is indeed some increase in the error rates at the edges of DeepLabV2, although it suggests that it is only noticeably detrimental for a small set of pixels that are closest to the edge. Therefore, if label accuracy is most important, it may be desirable to clip the 20-40 pixels that are closest to the edges of label patches. This approach introduces additional computational costs, but the proposed approach in Section VII.A (to increase the input-patch size) substantially mitigates this cost, and therefore we generally recommend clipping.

## VII. CONCLUSIONS

This work is focused on the problem of stitching - a strategy for processing large data that cannot be stored in its entirety on modern hardware (GPUs), and therefore it must be processed in a piecemeal fashion. A generic stitching approach is illustrated in Fig. 1, but there are several variants in the literature, and they are usually employed with little motivation or with little provided experimental support.

In this work we address the absence of investigation on stitching in the literature. We provide a survey of existing approaches and opinion on this topic. Based on this survey, we find that all variants of stitching are essentially motivated by *translational variance* in modern segmentation networks. We explain the two likely causes of translational variance, including experimental evidence of their existence and impact.

Based on our investigations we recommend a stitching strategy with the following guidelines:

- Enlarge the network's input-size *only during label inference*.
- Clip the edges of the output label images to remove higher-error labels
- There is little benefit to averaging labels from translated input, if recommendation (1) is followed. If label accuracy is a priority, then averaging labels from many (10 or more) translated input images can yield small, but consistent, accuracy improvements.

The first recommendation is particularly important – we show that it substantially reduces computation time, while also providing modest improvements in label accuracy. The proposed stitching approach was employed in our winning entry to the recent INRIA building labeling competition [7], [15], and is an extension of our preliminary work [13].

We note that, although our segmentation datasets are both focused on remote sensing applications, we believe our major conclusions should not depend upon the dataset or problem, and should apply to other application domains.

ACKNOWLEDGEMENTS

We thank the NVIDIA corporation for donating a GPU for this work, and the NSF XSEDE computational environment and the Duke Compute Clusters for providing computing resources. We would also like to thank the Duke University Energy Initiative for supporting our work.

APPENDIX I

Appendix I provides a sketch of the proof for the relationship between translational equivariance and non-unary-strided network layers in Section VI.B. To begin, let $\delta_1$ refer to the stride of a network layer $f_1(\cdot)$ that is operating (without loss of generality) on a 1-dimensional input $I'(j')$. So, $f_1(\cdot)$ could be a convolutional filtering operation or a pooling operation, and $I'(j')$ could be the input imagery, or the output of some intermediate layer. If we apply $f_1(\cdot)$ to $I'$, we have a new image $I''$ and its new coordinate system, denoted by $j''$. Furthermore, the relationship between $j''$ and $j'$ is given by equation (1) in Section V.B.

Based on the results discussed in Section V.B and illustrated in Fig. 5, we found that translations that are multiples of $\delta_1$ will result in translational equivariance. We can write this equivariance relationship as

$$I''(j'' + k) = f_1(I'(j' + \delta_1 k)) \quad \forall k \in \mathbb{N} \quad (3)$$

This simply states that translating $I'$ by factors of $\delta_1$ results in a translated, but otherwise equal, version of $I''$. Conversely, this will not be true for alternative translations of the input.

Let us now assume that $I'$ is actually the output of a previous function, $f_0$ with a stride of $\delta_0$ being applied to preceding input data, denoted $I(j)$. We want to know what kinds of translations in $I$ will satisfy the constraint in (3) for equivariance. To achieve this, we can use the relationship in equation (1) to map the permissible equivariant translations in (3) into the coordinate system of the input image $I$. This is given by

$$j = g^{-1}(j' + \delta_1 k) = \delta_1 j' + \delta_1 \delta_0 k \quad (4)$$

This equation suggests that translations in $I$ that are factors of $\delta_1 \delta_0 k$ will satisfy equation (3), ensuring that $I''$ is equivariant with respect to translations in $I$. Note that we can ignore $\delta_1 j'$ because it is a fixed arbitrary translation of $I$, and in practice this can be any value. Notice that maintaining the equivariance constraint for equation (3), also satisfies the constraint for translational equivariance between $I$ and $I'$. More precisely, if we let $k' = \delta_1 k$ then $k' \in \mathbb{N}$, satisfies the equivariance constraint (not explicitly written, but similar to (3)) with respect to $I'$.

Extending this result inductively (i.e., supposing there are more preceding layers in the network) yields equation (2) in the text. If all non-unary strides are the same, then we find that strides of $\delta^N$ in a network's input imagery should yield translation equivariant outputs from the it, where $N$ is the number of (not necessarily consecutive) layers.